\documentclass[sigconf]{acmart}
\AtBeginDocument{%
  }

\acmConference[KDD '25]{Proceedings of the 31st ACM SIGKDD Conference on Knowledge Discovery and Data Mining}{August 3--7, 2025}{Toronto, ON, Canada}
\usepackage{amsmath,amsfonts,bm}

\def\eqref#1{equation~\ref{#1}}
\def\1{\bm{1}}

\DeclareMathAlphabet{\mathsfit}{\encodingdefault}{\sfdefault}{m}{sl}
\SetMathAlphabet{\mathsfit}{bold}{\encodingdefault}{\sfdefault}{bx}{n}

\newcommand{\R}{\mathbb{R}}

\usepackage{hyperref}
\usepackage{url}
\usepackage{color}
\usepackage{wrapfig}
\usepackage{microtype}
\usepackage{graphicx}
\usepackage{booktabs} % for professional tables
\usepackage{multirow}
\usepackage{algorithm}
\usepackage{algorithmic}
\usepackage{amsmath}
\usepackage{mathtools}
\usepackage{amsthm}
\usepackage{mathtools}
\usepackage{bm,bbm}
\usepackage{amsfonts}
\usepackage{caption} 
\usepackage{placeins}

\theoremstyle{plain}
\newtheorem{theorem}{Theorem}[section]

\theoremstyle{definition}

\theoremstyle{remark}

\usepackage{amsmath}
    
\usepackage{amssymb}
    
\usepackage{mathtools}
\usepackage{amsthm}

\usepackage[capitalize,noabbrev]{cleveref}

\theoremstyle{plain}
\theoremstyle{definition}
\theoremstyle{remark}
\usepackage[textsize=tiny]{todonotes}
\usepackage[utf8]{inputenc}
\DeclareUnicodeCharacter{2000}{ }

\def \R {\mathbb{R}}

\def \Q {\mathcal{Q}}
\def \W {\mathcal{W}}
\def \N {\mathcal{N}}

\def \M {\mathcal{M}}

\begin{document}

\title{Does Vector Quantization Fail in Spatio-Temporal Forecasting?\\Exploring a Differentiable Sparse Soft-Vector Quantization Approach}

\author{Chao Chen}
\authornote{The first two authors contributed equally to this research.}
\email{ccchaochen@csu.edu.cn}
\affiliation{%
  \institution{Central South University}
  %\country{China}
  % \streetaddress{P.O. Box 1212}
  \city{Changsha}
  \country{China}
  % \postcode{}
}

\author{Tian Zhou}
\authornotemark[1]
% \authornote{Authors contributed equally to this research.}
% \authornote{Corresponding authors.}
% \authornotemark[2]
\email{tian.zt@alibaba-inc.com}
\affiliation{%
  \institution{DAMO Academy, Alibaba Group}
  %\country{China}
  % \streetaddress{P.O. Box 1212}
  \city{Hangzhou}
  \country{China}
  \postcode{}
}

\author{YanJun Zhao}
% \authornote{Corresponding authors.}
% \authornotemark[2]
\email{xiangyan.zyj@alibaba-inc.com}
\affiliation{%
  \institution{DAMO Academy, Alibaba Group}
  %\country{China}
  % \streetaddress{P.O. Box 1212}
  \city{Hangzhou}
  \country{China}
  \postcode{}
}

\author{Hui Liu}
% \authornote{Corresponding authors.}
% \authornotemark[2]
\email{csuliuhui@csu.edu.cn}
\affiliation{%
  \institution{Central South University}
  %\country{China}
  % \streetaddress{P.O. Box 1212}
  \city{Changsha}
  \country{China}
  \postcode{}
}

\author{Rong Jin}
% \authornote{Corresponding authors.}
% \authornotemark[2]
\email{rongjinemail@gmail.com}

\affiliation{%
  \institution{DAMO Academy, Alibaba Group}
  %\country{China}
  % \streetaddress{P.O. Box 1212}
  \city{Bellevue}
  \country{USA}
}

\author{Liang Sun}
\email{liang.sun@alibaba-inc.com}
\affiliation{%
  \institution{DAMO Academy, Alibaba Group}
  \city{Bellevue}
  \country{USA}
  % \streetaddress{P.O. Box 1212}
  % \city{Hangzhou}
  %\country{China}
}

\renewcommand{\shortauthors}{Chao Chen, Tian Zhou, YanJun Zhao, Hui Liu, Rong Jin, Liang Sun}

\begin{abstract}
%Spatio-temporal forecasting is crucial in various fields and relies on a careful balance between identifying subtle patterns and filtering out noise. Although clustering-based vector quantization methods are extensively applied in image generation tasks to enhance quality and filter noise, surprisingly, they fail to boost the accuracy of spatio-temporal forecasting. In response, we introduce Sparse Soft-Vector Quantization (SVQ), a groundbreaking technique with a strong foundation in theory and practice, proven to surpass traditional vector quantization methods in spatio-temporal forecasting. This approach preserves critical details from the original vectors using a regression model while filtering out noise via sparse design. Moreover, we approximate the sparse regression process using a blend of a two-layer MLP and an extensive codebook. This strategy not only substantially cuts down on computational costs but also grants SVQ differentiability and training simplicity, resulting in a notable enhancement of performance. Our empirical studies on five spatio-temporal benchmark datasets demonstrate that SVQ achieves state-of-the-art results. Specifically, on the WeatherBench-S temperature dataset, SVQ improves the top baseline by 7.9\%. In video prediction benchmarks—Human3.6M, KTH, and KittiCaltech—it reduces MAE by an average of 9.4\% and improves image quality by 17.3\% (LPIPS). Code is publicly available at \url{https://anonymous.4open.science/r/SVQ-Forecasting}.

Spatio-temporal forecasting is crucial in various fields and requires a careful balance between identifying subtle patterns and filtering out noise. Vector quantization (VQ) appears well-suited for this purpose, as it quantizes input vectors into a set of codebook vectors or patterns. Although VQ has shown promise in various computer vision tasks, it surprisingly falls short in enhancing the accuracy of spatio-temporal forecasting. We attribute this to two main issues: inaccurate optimization due to non-differentiability and limited representation power in hard-VQ. To tackle these challenges, we introduce Differentiable Sparse Soft-Vector Quantization (SVQ), the first VQ method to enhance spatio-temporal forecasting. SVQ balances detail preservation with noise reduction, offering full differentiability and a solid foundation in sparse regression. Our approach employs a two-layer MLP and an extensive codebook to streamline the sparse regression process, significantly cutting computational costs while simplifying training and improving performance. Empirical studies on five spatio-temporal benchmark datasets show SVQ achieves state-of-the-art results, including a 7.9\% improvement on the WeatherBench-S temperature dataset and an average mean absolute error reduction of 9.4\% in video prediction benchmarks (Human3.6M, KTH, and KittiCaltech), along with a 17.3\% enhancement in image quality (LPIPS). Code is publicly available at \url{https://github.com/Pachark/SVQ-Forecasting}.

\end{abstract}


  


\maketitle

\section{Introduction}
Spatio-temporal forecasting is pivotal in numerous domains ranging from environmental monitoring to urban planning, where precisely predicting future dynamics is crucial. Among various methodologies explored in deep learning, Vector Quantization (VQ) has distinguished itself primarily in computer vision tasks, showcasing its ability to compress high-dimensional vectors into a compact, discrete form that maintains significant fidelity to the original information. Historically rooted in signal processing, VQ's breakthrough came with its application in image processing advancements like the Vector Quantized-Variational AutoEncoder (VQ-VAE)~\cite{vqvae}, which set a precedent in generating high-quality images by learning efficient representations of complex distributions.

\begin{figure}
\centering
% \vskip -0.1in
\includegraphics[width=0.45\textwidth]{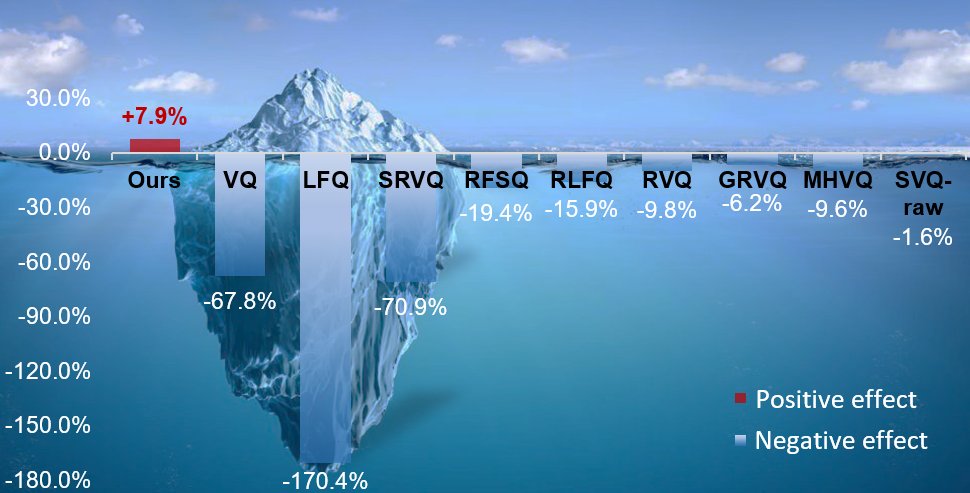}
\vskip -0.1in
\caption{\textbf{Limitations of VQ in spatio-temporal forecasting:} An experiment study evaluating MSE improvement percentage on the WeatherBench-S temperature dataset.}
% \vskip -0.10in
\label{fig:VQ_fail}
\end{figure}

While VQ has proven effective and has become a nearly default approach in generation tasks, its potential in spatio-temporal forecasting remains underexplored. In spatio-temporal forecasting, data are considered including patterns, details, and noise. VQ offers the capability to capture recurring patterns through a discrete codebook, while simultaneously filtering out noise and irrelevant details that are unlikely to recur. Given its noise reduction properties and the similarities between image/video generation and spatio-temporal forecasting, one might reasonably expect VQ to yield performance gains in the latter. However, our review of existing studies reveals that few have successfully leveraged VQ to enhance forecasting accuracy. Our empirical evaluation of recent VQ methods shows that all fell short of expectations, often degrading the performance of baseline forecasting models rather than providing the expected improvements, as illustrated in Figure \ref{fig:VQ_fail} and detailed in Table \ref{tab:ablation_vq_method}. They consistently exhibited a negative impact on final forecasting metrics such as mean squared error (MSE).

We believe that the similarity between image/video generation and spatio-temporal forecasting underscores the potential of VQ, but the problem lies in the inherent dynamic nature of spatio-temporal data. The complex temporal evolutions and spatial distributions introduce a level of complexity that traditional VQ methods are not equipped to handle. Specifically, the unsatisfactory results of traditional VQ methods are caused by two reasons: 

\textbf{Inaccurate model optimization caused by non-differentiability.} The discrete nature of the quantization step prevents gradients from being directly passed through this operation. VQ methods typically employ the straight-through (or stop-gradient) estimator, as described in VQVAE~\cite{vqvae}. This estimator approximates the gradient by copying gradients from the quantized outputs to the input vectors, introducing errors in the optimization process.

\textbf{Limited representation power of hard-VQ.} The rapid evolution of spatiotemporal data over time, makes learning transformations from historical inputs to future outputs challenging. Hard-VQ struggles to replace the latent vector with the nearest codebook vector as the latent changes quickly, which limits the modeling of detailed spatio-temporal dynamics required for forecasting.

In response to these limitations, this work presents Differentiable Sparse Soft-Vector Quantization (SVQ), designed to strike a balance between noise reduction and detail preservation for spatio-temporal forecasting tasks. We solve the aforementioned challenges by:
\begin{itemize}
   \item Introduced a differentiable VQ mechanism to enable efficient gradient propagation. This involves using a two-layer MLP to approximate sparse regression, maintaining differentiability and enhancing accuracy. By generating regression coefficients through nonlinear projections and deriving quantized outputs using a codebook matrix, the approach facilitates gradient flow from outputs to inputs, addressing computational challenges in traditional VQ.
   \item Developed Soft-VQ, a method that integrates sparse regression with a large codebook, allowing multiple vector allocations for each input. Compared to hard-VQ, SVQ effectively captures complex patterns while enhancing noise filtration. It preserves diverse information from inputs, better capturing dynamic features in spatio-temporal data. Remarkably, SVQ can utilize a static, randomly initialized codebook without performance degradation, thus reducing learning parameters and demonstrating efficiency and robustness.
\end{itemize}

% The efficiency of VQ in summarizing static information contrasts sharply with the demands of modeling detailed spatio-temporal dynamics, where high temporal resolution and the complex interactions among spatial elements over time are crucial.

% \begin{figure*}
% \centering
% \includegraphics[width=1\linewidth]{graphs/codebook_tsne_allocation.png}
% \vskip -0.1in
% \caption{{\bf Hard-VQ vs Soft-VQ:} \small 
% Using t-SNE, we visualize the codebook vectors on WeatherBench-S temperature dataset with the same codebook size (1024). We provide examples of codebook allocation, where hard-VQ typically assigns each input vector to a single codebook vector as described in VQ-VAE~\cite{vqvae}, while soft-VQ assigns each input vector to multiple codebook vectors.}
% \vskip -0.1in
% \label{fig:codebook_tsne_vq}
% \end{figure*}

Through extensive evaluation on a variety of real-world datasets, SVQ emerges as the first VQ method to achieve significant enhancements in spatio-temporal forecasting tasks. Notably, SVQ surpassed the leading model in WeatherBench-S temperature forecasting benchmark, achieving a 7.9\% reduction in MSE. In video prediction tasks-Human3.6M, KTH, and KittiCaltech, SVQ systematically reduced the Mean Absolute Error (MAE) by 9.4\%, and significantly improved perceptual quality, as indicated by a 17.3\% reduction in the learned perceptual image patch similarity (LPIPS) score. These results underscore SVQ's remarkable capability across a wide range of spatio-temporal forecasting scenarios.

\section{Related Work}
Here, we present a brief overview of vector quantization, the lineage of sparse coding techniques, and the latest developments in spatio-temporal forecasting models.

\textit{Vector Quantization and Sparse Coding.}\quad Instead of using continuous latent, VQ-VAE~\cite{vqvae}, a seminal work, incorporates vector quantization to learn discrete latent representations, typically assigning each vector to the nearest code in a codebook. Subsequent enhancements include Residual VQ~\cite{residualvq}, which quantizes the residuals recursively, and Multi-headed VQ~\cite{multiheadvq}, which adopts multiple heads for each vector. While these methods are effective, they often rely on a relatively small number of codes to represent the original vectors. To address this, SCVAE~\cite{scvae} employs sparse coding, allowing vectors to be represented through sparse linear combinations of multiple codes, and achieves end-to-end training via the Learnable Iterative Shrinkage Thresholding Algorithm (LISTA)~\cite{lista}. However, a significant drawback of the sparse coding method using LISTA (referred to as SVQ-raw here) is its high computational complexity, which scales quadratically with codebook size.

Building on these insights and limitations, our work proposes a soft-VQ method applicable to spatio-temporal forecasting tasks. Although it is closely related to a simultaneous research work~\cite{Tschannen2023GIVTGI}, which employs an infinite cookbook with a linear layer for continuous vector quantization applied in image generation, our approach is largely inspired by sparse regression, as clearly evidenced by our analysis. Specifically, our work focuses on the challenges arising from spatio-temporal forecasting, providing a strong theoretical foundation and effectively addressing the challenges. 

\textit{Spatio-Temporal Forecasting Models.}\quad Recent advancements in spatio-temporal forecasting have highlighted a shift from recurrent to non-recurrent frameworks. Despite the forecasting capabilities of models like ConvLSTM~\cite{convlstm}, PredRNN~\cite{predrnn}, and PredRNNV2\cite{predrnnv2}, this shift is largely due to the high computational demands of sequential processing in recurrent models. Non-recurrent models, such as MMVP~\cite{mmvp} and the SimVP family~\cite{SimVP,simvpv2}, have become benchmarks in video prediction by decoupling spatial and temporal learning through an efficient encoder-translator-decoder structure. This transition is further enhanced by innovative features like visual attention in TAU~\cite{tau} and MetaFormers in OpenSTL~\cite{openstl}, showcasing the continuous improvements towards more effective forecasting solutions. Our proposed method is designed for seamless integration as a plugin with the majority of these spatio-temporal forecasting models.

%\textcolor{red}{We acknowledge the simultaneous research ~\cite{Tschannen2023GIVTGI}, which employs an infinite cookbook with a linear layer for continuous vector quantization in traditional image generation. Independently, our work parallels this by applying a similar conceptual framework to spatiotemporal forecasting, adopting a distinct sparse regression perspective. The coinciding development of these studies underscores the community's burgeoning interest in continuous regression-based VQ methods. Together, our contributions offer valuable, complementary insights to the field.}

% To the best of our knowledge, sparse vector quantization has not been applied to spatiotemporal forecasting yet. We hypothesize that sparse vector quantization ensures a good generalized performance for spatiotemporal forecasting.
\section{Differentiable Sparse Soft-Vector Quantization (SVQ)}

We first present the mathematical foundation of differentiable sparse soft-vector quantization. Our proposed method effectively solves the optimization problem in a differentiable manner, and the theoretical analysis of codebook utilization demonstrates its strong representational capacity. We then describe its detailed implementation within a spatio-temporal forecasting model.

\subsection{Vector Quantization by Sparse Regression}
\label{sec:sparse_regression}
Let $\{z_i \in \R^d\}_{i=1}^m$ be the set of codes. A typical vector quantization method assigns a data point $x \in \R^d$ to the nearest code in $\{z_i\}_{i=1}^m$. The main problem with such an approach is that a significant part of the information in $x$ will be lost due to quantization. Sparse regression turns the code assignment problem into an optimization problem
\begin{eqnarray}
w = \mathop{\arg\min}_{w \in \R_+^m} \frac{1}{2}\left|x - \sum_{i=1}^m w_i z_i\right|^2 + \lambda |w|_1,
\label{eqn:1}
\end{eqnarray}

% \end{figure}

\noindent where $w = (w_1, \ldots, w_m) \in \R_+^m$ is the weight for combining codes $\{z_i\}_{i=1}^m$ to approximate $x$. $\lambda$ refers to the regularization parameter. By introducing $L_1$ regularizer in the optimization problem, we effectively enforce $x$ to be associated with a small number of codes. Compared to classic VQ methods where codes have to be learned through clustering, according to \cite{Random-projection}, it is sufficient to use randomly sampled vectors as codes as long as its number is large enough, thus avoiding the need of computing and adjusting codes. The theoretical guarantee of sparse regression is closely related to the property of subspace clustering, as revealed in \cref{theorem}.

\begin{figure}[h]
\centering
\includegraphics[width=0.4\textwidth]{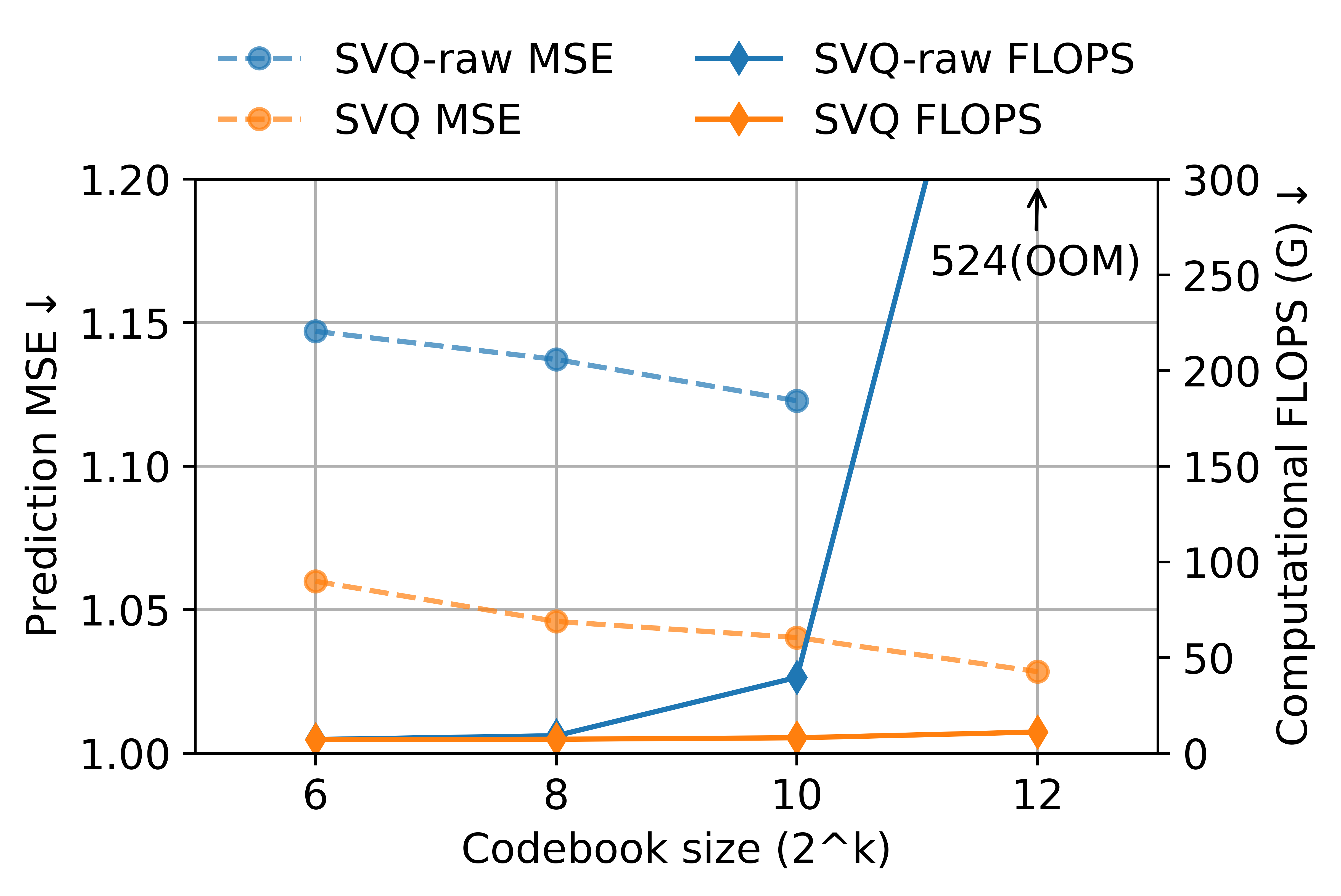}
\vskip -0.1in
\caption{{\bf Effect of SVQ approximation:} Floating point operations per second (FLOPs) and mean squared error (MSE) on WeatherBench-S temperature dataset with SVQ-raw and SVQ. The computational complexity of SVQ-raw increases quadratically with the size of codebook, making it suffer from out-of-memory (OOM) issue when scaling codebook size up to $2^{12}$.}
\label{fig:compare_lista_mlp}
\end{figure}

As shown in Figure \ref{fig:compare_lista_mlp}, the obvious downside of sparse regression for VQ is its high computational cost, as it needs to solve the optimization problem in (\ref{eqn:1}) for EVERY data point. Below, we will show that sparse regression can be approximated by a two-layer MLP and a randomly fixed or learnable matrix, making it computationally attractive.

% \begin{figure}

\begin{figure*}[hbt]
  \centering
  \includegraphics[width=1\linewidth]{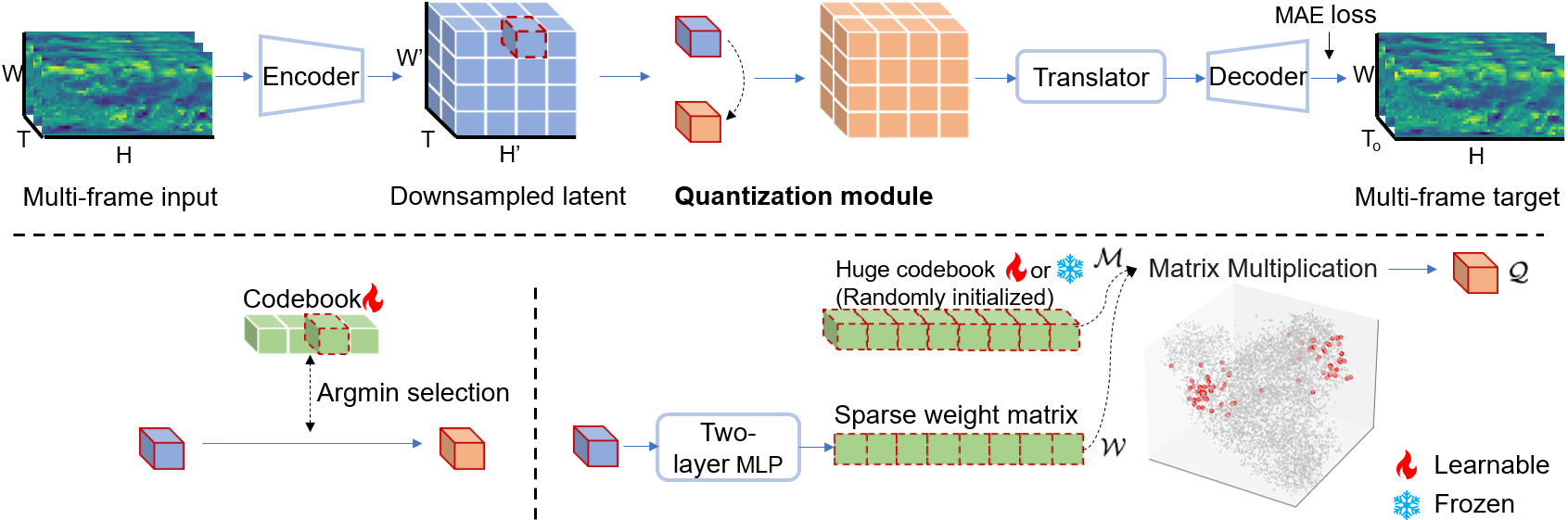}
  \caption{{\bf Top: Architecture of backbone model and the proposed quantization module.} The encoder, translator, decoder are inherited from SimVP~\cite{simvpv2}. A quantization module is added between the encoder and translator to effectively ensure a good generalized performance. {\bf Bottom: Quantization process of traditional VQ (Left) and our proposed SVQ (Right)}. In contrast, SVQ select multiple codes (\textcolor{red}{red} dots) from a huge codebook (\textcolor{gray}{gray} dots), and the codebook can be either learnable or frozen.}
  \label{fig:architecture}
\end{figure*}

% \begin{figure}[h]
%   \centering
%   \includegraphics[width=0.5\linewidth]{graphs/compare_lista_mlp.png}
%   \caption{\small {\bf Effect of SVQ approximation:} Computational cost (FLOPS) and prediction error (MSE) on WeatherBench-S temperature dataset with SVQ-raw and SVQ. The computational complexity of SVQ-raw increases quadratically with the size of codebook, making it suffer from out-of-memory (OOM) issue when scaling codebook size up to $2^{12}$.}
%   \label{fig:compare_lista_mlp}
% \end{figure}

To solve the optimization problem (\ref{eqn:1}), we consider the composite optimization method whose iteration is given as follows
\begin{eqnarray}
w'_{t+1} & = & w_t - \eta Z^{\top}(Zw_t - x), \\ \left[w_{t+1}\right]_i & = &\mbox{sgn}\left(\left[w'_{t+1}\right]_i\right) \left(\left|\left[w'_{t+1}\right]_i\right| - \lambda \eta\right)_+,
\end{eqnarray}
where $Z = (z_1, \ldots, z_m)$, $[z]_i$ is the $i$th element of vector $z$, $\mbox{sgn}$ denotes the sign function, and $(a)_+$ outputs $0$ if $a < 0$ and $a$ otherwise. We consider the first step of the iteration where $w_0 = 0$ and have $w = \eta \mbox{sgn}\left(Z^{\top} x - \lambda \mathbf{1})\right[Z^{\top} x - \lambda \mathbf{1}]_+ = \eta \mbox{sgn}\left(Z^Tx - \lambda\right)\sigma\left(Z^Tx - \lambda\right)$ and the resulting output for $x$ is given as
\begin{eqnarray}
x' = \sum_{i=1}^m w_i z_i = \eta Z sgn\left(Z^Tx - \lambda\right)\sigma\left(Z^Tx - \lambda\right). 
\end{eqnarray}
By generalizing $\eta Z$ into another matrix $B$, we have output vector $x'$ exactly expressed as a matrix and a two-layer MLP over $x$. We finally note that although it is convenient to form the codebook by randomly sampling vectors, we found empirically that tuning codebook does bring slight additional gains in some cases. 

\subsection{Efficient Utilization of Cookbook Using Sparse Regression}
\label{sec:comp_svq_cluster}
To understand the difference between sparse regression-based quantization scheme and clustering-based quantization scheme, we measure the number of codes required to approximate any vector within a unit ball $\mathcal{B}$ with error less than $\delta$. This number is denoted by $T(\mathcal{B}, \delta)$. Intriguingly, as the theorem below reveals, using sparse regression allows $T(\mathcal{B}, \delta)$ to be significantly reduced from $O(1/\delta^d)$ to $O(1/\delta^p)$, where $p \ll d$ for high-dimensional vectors.

\begin{theorem}
\label{theorem}
For the clustering-based method, $T(\mathcal{B}, \delta)$ is at least $1/\delta^d$. In contrast, for sparse regression, $T(\mathcal{B},\delta)$ can be formulated as $(4d/\delta)^p$, where 
\begin{eqnarray}
p \geq \max\left(3, \frac{\log(4/\delta)}{\log\log(2d/\varepsilon)}\right),
\end{eqnarray}
given that the number of non-zero elements utilized by sparse regression is at least
\begin{eqnarray}
\frac{4d}{\delta \left(\log C + p\log (4d) - (p+1)\log \delta \right)}.
\end{eqnarray}
\end{theorem}

\begin{proof}
\renewcommand{\qedsymbol}{}
To estimate $T(\mathcal{B}, \delta)$ for the clustering method, we consider the covering number for a unit ball $\mathcal{B}$ which necessitates at least $1/\delta^d$ code vectors to approximate any vector within an acceptable error margin of $\delta$. With $U = (u_1, \ldots, u_m)$ where $u_k \sim \N(0, I_d/m)$, and $g \in \Delta_s$ an $s$-sparse unit vector, we discern:
\begin{eqnarray}
\Pr\left(\|UU^{\top} - I\|_2 \geq \gamma \right) \leq 2d\exp\left(-\frac{m\gamma^2}{3d}\right),
\end{eqnarray}
which implies
\begin{eqnarray}
\|UU^{\top} - I\|_2 \leq \Delta := \sqrt{\frac{d}{m}\log\frac{2d}{\varepsilon}}
\end{eqnarray}
with probability at least $1 - \varepsilon$. Therefore, $\|g' - g\|_2 \geq (1+\Delta)^{-1} \|Ug - Ug'\|_2$, where $g'$denotes a covering vector. Since the $s$-sparse unit vector covering number is bounded by $(Cm/s\delta)^s$, we establish:
\begin{eqnarray}
\left(\frac{Cm}{s\delta}\right)^s \geq \left(1 + \frac{2}{\delta}\right)^d(1+\Delta)^d,
\end{eqnarray}
Setting $m = (4d/\delta)^p$ yields
\begin{eqnarray}
(1+\Delta)^d \leq \exp(d\Delta) \leq e,
\end{eqnarray}
therefore, $s\log(C'm/\delta) \geq d(2/\delta + \log(1+\Delta))$, where $C' = Ce$. As long as $s \geq \frac{4d}{\delta(\log C + p\log(4d) - (p+1)\log\delta)}$, it follows that $s\log s \leq 2d/\delta$, affirming that $m \geq (4d/\delta)^p$.
\end{proof}

\subsection{Spatio-Temporal Forecasting Model Enhanced by Quantization}

\textit{\indent Architecture of backbone model.}\quad As shown in Figure~\ref{fig:architecture}, SimVP~\cite{simvpv2} is employed as the backbone model, which encompasses an encoder for spatial feature extraction, a translator for temporal dependency learning, and a decoder for frame reconstruction. The quantization module is integrated between the encoder and translator. The input data is a 4D tensor $X \in \R^{H*W*T*C}$, representing height ($H$), width ($W$), time step ($T$), and channel ($C$). The encoder ${\rm En}$ condenses $X$ into downsampled latent representation ${\rm En}(X) \in \R^{H'*W'*T*C'}$, maintaining temporal dimensionality while altering spatial and channel dimensions. This latent space, composed of $H'*W'*T$ tokens, each represented by a $C'$-dimensional vector, undergoes vector quantization.

\textit{Quantization module.}\quad The SVQ comprises a two-layer MLP and an extensive codebook. The codebook is a randomly initialized matrix $\M \in \R^{N*C'}$, where $N$ denotes the size of codebook. To achieve automatic selection of codes, a weight matrix $\W \in \R^{H'*W'*T*N}$ is generated via nonlinear projection from the latent representation $\rm{En}(X)$. This projection is formally expressed as $\W = {\rm MLP}(\rm{En}(X))$, wherein the $\rm MLP$ comprises two linear layers and an intermediate ReLU activation function. The quantized output $\Q$ is then obtained by computing the matrix multiplication of weight matrix $\W$ and codebook matrix $\M$, a process that can be conceptualized as a selection operation as shown in Figure~\ref{fig:architecture}. To encourage sparsity within the generated weight matrix, we apply a Mean Absolute Error (MAE) loss to the output as a surrogate form of regularization.

\section{Experiments}
\label{sec:experiment}
We evaluate SVQ under the unified framework of OpenSTL~\cite{openstl}.

% The evaluation begins by assessing SVQ's ability to enhance various backbones. Subsequently, we compare the performance of the SimVP+SVQ model with state-of-the-art (SOTA) counterparts. This is followed by an investigation into the delicate balance between detail preservation and noise reduction, and an experiment with artificial noise injection to show SVQ's effectiveness in mitigating noise. Finally, we perform a series of ablation studies. Additionally, we provide supplementary experiments to understand the effect of SVQ on latent representation, and compare different VQ methods in Appendix \ref{sec:appendix_latent}, and \ref{sec:additional_vq}, respectively.

\textit{Datasets.}\quad We conduct experiments on five real-world spatio-temporal forecasting tasks, including weather (WeatherBench~\cite{weatherbench}), traffic flow (TaxiBJ~\cite{taxibj}), human pose dynamics (Human3.6M~\cite{human3.6m}), driving scenes (KittiCaltech~\cite{kitti, caltech}), and human actions (KTH Action~\cite{kth}). Details about datasets are provided in Appendix \ref{sec:appendix_dataset}.

% A summary of dataset statistics is provided in Table \ref{tab:dataset}. More details about datasets are described in Appendix 
% \ref{sec:appendix_dataset}.
% \vskip -0.1in
% \input{tables/tab_dataset.tex}
% \vskip -0.1in

\textit{Experimental details.}\quad We found SVQ to be quite robust to codebook size, as its performance remains consistently strong when using a sufficiently large codebook. Therefore, we fix the codebook size at 10,000 for WeatherBench, TaxiBJ, and Human3.6M, and at 6,000 for KittiCaltech and KTH. The hidden dimension of nonlinear projection layer is fixed at 128. Experiments are conducted on 1 or 4 NVIDIA V100 32GB GPUs, with a total batch size of 16, and early stopped with a patience of 10. Details about backbone architectures and computational costs are described in Appendix \ref{sec:appendix_architecture} and \ref{sec:appendix_cost}.

\textit{Metrics.}\quad Forecasting accuracy is evaluated using mean squared error (MSE) and mean absolute error (MAE), while the image quality of predicted frames is assessed using structural similarity index measure (SSIM)~\cite{ssim}, peak signal-to-noise ratio (PSNR), and learned perceptual image patch similarity (LPIPS)~\cite{lpips}. 

\textit{Baselines.}\quad The comparison baselines include: 1) Non-recurrent models such as SimVP~\cite{simvpv2} and TAU~\cite{tau}; 2) Recurrent-based models such as ConvLSTM~\cite{convlstm}, PredNet~\cite{prednet}, PredRNN~\cite{predrnn}, MIM~\cite{mim}, E3D-LSTM~\cite{e3dlstm}, PhyDNet~\cite{phydnet}, MAU~\cite{mau}, PredRNNv2~\cite{predrnnv2}, and DMVFN~\cite{dmvfn}. Baseline results are copied from OpenSTL~\cite{openstl}. For SimVP, we report the best MetaFormer variant on each dataset, detailed in Appendix \ref{sec:appendix_architecture}. It is worth noting that while some diffusion-based methods~\cite{feng2023diffpose,chen2023seine,gupta2024photorealistic} have shown strong performance in video generation, our focus is on spatiotemporal forecasting, which is typically evaluated using MSE and MAE rather than image authenticity. Therefore, we did not include them in our baselines. Further exploration is needed to understand their potential in forecasting settings, given the complex noise-signal interactions.

\begin{table*}[t]
\centering
\caption{{\bf WeatherBench results:} Performance comparison for SVQ module and baseline models on WeatherBench. WeatherBench-S is single-variable, one-hour interval forecasting setup trained on data from 2010-2015. WeatherBench-M is a multi-variable, six-hour interval forecasting setup designed for broader applications, trained on data from 1979-2015. Both datasets are validated on 2016 and tested on 2017-2018. The best and second-best results are highlighted by \textbf{bold} and \underline{underlined}.}
\begin{center}
\begin{small}
\vskip -0.1in
\scalebox{1.0}{
% \begin{sc}
\begin{tabular}{c|cc|cc|cc|cc|cc}
\toprule
\multirow{2}{*}{Dataset}&\multicolumn{2}{c|}{Variable}&\multicolumn{2}{c|}{Temperature}&\multicolumn{2}{c|}{Humidity}&\multicolumn{2}{c|}{Wind Component}&\multicolumn{2}{c}{Total Cloud Cover}\\
% \midrule
\cmidrule{2-11}
 &\multicolumn{2}{c|}{Model}&MSE$\downarrow$ & MAE$\downarrow$ & MSE$\downarrow$ & MAE$\downarrow$ &MSE$\downarrow$ & MAE$\downarrow$ & MSE$\downarrow$ & MAE$\downarrow$ \\
\midrule

\multirow{10}{*}{WeatherBench-S}&\multicolumn{2}{c|}{ConvLSTM\cite{convlstm}}& 1.521 & 0.7949 & 35.146 & 4.012 & 1.8976 & 0.9215 & 0.0494  & 0.1542  \\
&\multicolumn{2}{c|}{E3D-LSTM\cite{e3dlstm}}& 1.592 & 0.8059 & 36.534 & 4.100 & 2.4111 & 1.0342 & 0.0573  & 0.1529  \\
&\multicolumn{2}{c|}{PredRNN\cite{predrnn}}& 1.331 & 0.7246 & 37.611 & 4.096 & 1.8810 & 0.9068 & 0.0550  & 0.1588  \\
&\multicolumn{2}{c|}{MIM\cite{mim}}& 1.784 & 0.8716 & 36.534 & 4.100  & 3.1399 & 1.1837 & 0.0573  & 0.1529  \\
&\multicolumn{2}{c|}{MAU\cite{mau}}& 1.251  & 0.7036 & 34.529  & 4.004 & 1.9001 & 0.9194 & 0.0496  & 0.1516  \\
&\multicolumn{2}{c|}{PredRNN++\cite{predrnn++}}& 1.634 & 0.7883 & 35.146 & 4.012 & 1.8727 & 0.9019 & 0.0547 & 0.1543 \\
&\multicolumn{2}{c|}{PredRNN.V2\cite{predrnnv2}}& 1.545 & 0.7986 & 36.508 & 4.087 & 2.0072 & 0.9413 & 0.0505  & 0.1587  \\
&\multicolumn{2}{c|}{TAU\cite{tau}}& 1.162 & 0.6707 & 31.831 & 3.818 & 1.5925 & 0.8426 & 0.0472 & \underline{0.1460}  \\
&\multicolumn{2}{c|}{SimVP (w/o VQ)\cite{simvpv2}}& 1.105 & 0.6567 & 31.332 & 3.776 & 1.4996 & 0.8145 & 0.0466  & 0.1469  \\
&\multicolumn{2}{c|}{\textbf{SimVP+SVQ (Frozen codebook)}} &\underline{1.023}& \underline{0.6131} & \underline{30.863} & \underline{3.661} & \underline{1.4337} &\underline{0.7861} & \textbf{0.0456} & \textbf{0.1456}\\
&\multicolumn{2}{c|}{\textbf{SimVP+SVQ (Learnable codebook)}} &\textbf{1.018} &\textbf{0.6109} &\textbf{30.611} &\textbf{3.657} &\textbf{1.4186}&\textbf{0.7858}&\underline{0.0458}&0.1463\\
&\multicolumn{2}{c|}{\textbf{Improvement}} & $\uparrow$\textbf{7.9\%}& $\uparrow$\textbf{7.0\%} & $\uparrow$\textbf{2.3\%} & $\uparrow$\textbf{3.2\%} & $\uparrow$\textbf{5.4\%} & $\uparrow$\textbf{3.5\%} & $\uparrow$\textbf{2.1\%} & $\uparrow$\textbf{0.9\%}\\
\midrule
\multirow{10}{*}{WeatherBench-M}&\multicolumn{2}{c|}{Variable}&\multicolumn{2}{c|}{Temperature}&\multicolumn{2}{c|}{Humidity}&\multicolumn{2}{c|}{Wind U Component}&\multicolumn{2}{c}{Wind V Component}\\
\cmidrule{2-11}
&\multicolumn{2}{c|}{ConvLSTM\cite{convlstm}}& 6.303 & 1.7695 & 368.15 & 13.490 & 30.002 & 3.8923 & 30.789 & 3.8238 \\
&\multicolumn{2}{c|}{PredRNN\cite{predrnn}}&5.596 & 1.6411 & 354.57 & 13.169 & 27.484 & 3.6776 & 28.973 & 3.6617 \\
&\multicolumn{2}{c|}{MIM\cite{mim}}& 7.515 & 1.9650 & 408.24 & 14.658 & 35.586 & 4.2842 & 36.464 & 4.2066 \\
&\multicolumn{2}{c|}{MAU\cite{mau}}& 5.628 & 1.6810 & 363.36 & 13.503 & 27.582 & 3.7409 & 27.929 & 3.6700 \\
&\multicolumn{2}{c|}{PredRNN++\cite{predrnn++}}& 5.647 & 1.6433 & 363.15 & 13.246 & 28.396 & 3.7322 & 29.872 & 3.7067 \\
&\multicolumn{2}{c|}{PredRNN.V2\cite{predrnnv2}}& 6.307 & 1.7770 & 368.52 & 13.594 & 29.833 & 3.8870 & 31.119 & 3.8406 \\
&\multicolumn{2}{c|}{TAU\cite{tau}}& 4.904 & 1.5341 & \underline{342.63} & 12.801 & 24.719 & 3.5060 & 25.456 & 3.4723  \\
&\multicolumn{2}{c|}{SimVP (w/o VQ)\cite{simvpv2}} &4.833 & 1.5246 & \textbf{340.06} & 12.738 & 24.535 & 3.4882 & 25.232 & 3.4509 \\
&\multicolumn{2}{c|}{\textbf{SimVP+SVQ (Frozen codebook)}}& \textbf{4.427} & \textbf{1.4160} & 360.15 & \textbf{12.445} & \underline{23.915} & \underline{3.4078} & \textbf{24.968} & \underline{3.4117} \\
&\multicolumn{2}{c|}{\textbf{SimVP+SVQ (Learnable codebook)}} & \underline{4.433} & \underline{1.4164} & 360.53 & \underline{12.449} & \textbf{23.908} & \textbf{3.4060} & \underline{24.983} & \textbf{3.4095}\\
&\multicolumn{2}{c|}{\textbf{Improvement}} & $\uparrow$\textbf{8.4\%}& $\uparrow$\textbf{7.1\%} & $\downarrow$5.9\% & $\uparrow$\textbf{2.3\%} & $\uparrow$\textbf{2.6\%} & $\uparrow$\textbf{2.4\%} & $\uparrow$\textbf{1.0\%} & $\uparrow$\textbf{1.2\%}\\

\bottomrule
\end{tabular}
\label{tab:weatherbench}
}
% \end{sc}
\end{small}
\end{center}
\end{table*}
% \vskip -0.2in

\subsection{Benchmarks on various forecasting tasks}

\begin{table}
\centering
% \captionsetup{font=small} 
\caption{{\bf TaxiBJ results:} Performance comparison for SVQ module and baseline models on TaxiBJ. The best and second-best results are highlighted by \textbf{bold} and \underline{underlined}.}

\begin{center}
\begin{small}
\vskip -0.1in
% \begin{sc}
\scalebox{0.9}{
\begin{tabular}{cc|cccc}
\toprule
\multicolumn{2}{c|}{Model} &MSE$\downarrow$ & MAE$\downarrow$ & SSIM$\uparrow$& PSNR$\uparrow$\\
\midrule
\multicolumn{2}{c|}{ConvLSTM\cite{convlstm}}& 0.3358 & 15.32 & 0.9836 &39.45 \\
\multicolumn{2}{c|}{E3D-LSTM\cite{e3dlstm}}& 0.3427 & 14.98 & 0.9842 & 39.64\\
\multicolumn{2}{c|}{PhyDNet\cite{phydnet}}& 0.3622 & 15.53 & 0.9828 & 39.46\\
\multicolumn{2}{c|}{PredNet\cite{prednet}}& 0.3516 & 15.91 & 0.9828 & 39.29\\
\multicolumn{2}{c|}{PredRNN\cite{predrnn}}& 0.3194 & 15.31 & 0.9838 & 39.51\\
\multicolumn{2}{c|}{MIM\cite{mim}}& \underline{0.3110} & 14.96 & 0.9847 & 39.65\\
\multicolumn{2}{c|}{MAU\cite{mau}}& 0.3268 & 15.26 & 0.9834 & 39.52\\
\multicolumn{2}{c|}{DMVFN\cite{dmvfn}}& 3.3954 & 45.52 & 0.8321 & 31.14 \\
\multicolumn{2}{c|}{PredRNN++\cite{predrnn++}}& 0.3348 & 15.37 & 0.9834 & 39.47\\
\multicolumn{2}{c|}{PredRNN.V2\cite{predrnnv2}}& 0.3834 & 15.55 & 0.9826 & 39.49\\
\multicolumn{2}{c|}{TAU\cite{tau}}& \textbf{0.3108} 	&14.93 	&0.9848 	&39.74\\
\multicolumn{2}{c|}{SimVP (w/o VQ)\cite{simvpv2}}&0.3246 &15.03&0.9844&39.71 
\\
\multicolumn{2}{c|}{\textbf{SimVP+SVQ (Frozen codebook)}}& 0.3171  & \underline{14.68} & \underline{0.9848}  & \underline{39.83}  \\
\multicolumn{2}{c|}{\textbf{SimVP+SVQ (Learnable codebook)}}& 0.3191 & \textbf{14.64} & \textbf{0.9849} & \textbf{39.86}\\
\multicolumn{2}{c|}{\textbf{Improvement}}&$\uparrow$\textbf{1.7\%} &$\uparrow$\textbf{2.6\%}&$\uparrow$\textbf{0.1\%}&$\uparrow$\textbf{0.4\%} \\

\bottomrule
\end{tabular}
\label{tab:taxibj}
}
% \end{sc}
\end{small}
\end{center}
% \vskip -0.3in
%\vspace{-5mm}
\end{table}

\begin{table*}
\centering
\caption{{\bf Video prediction results:} Performance comparison for SVQ module and baseline models on Human3.6M, KTH, and KittiCaltech. The best and second-best results are highlighted by \textbf{bold} and \underline{underlined}.}
\begin{center}
\begin{small}
% \begin{sc}
\vskip -0.1in
\scalebox{0.90}{
\begin{tabular}{cc|cccc|cccc|cccc}
\toprule
\multicolumn{2}{c|}{Dataset} &\multicolumn{4}{c|}{Human3.6M}&\multicolumn{4}{c|}{KittiCaltech}&\multicolumn{4}{c}{KTH}\\
\midrule
\multicolumn{2}{c|}{Metric} &MAE$\downarrow$ & SSIM$\uparrow$& PSNR$\uparrow$ &LPIPS$\downarrow$&MAE$\downarrow$ & SSIM$\uparrow$& PSNR$\uparrow$ &LPIPS$\downarrow$&MAE$\downarrow$ & SSIM$\uparrow$& PSNR$\uparrow$ &LPIPS$\downarrow$\\
\midrule
\multicolumn{2}{c|}{ConvLSTM\cite{convlstm}}& 1583.3          & 0.9813          & 33.40 & 0.03557          & 1583.3          & 0.9345          & 27.46          & 0.08575     & 445.5          & 0.8977          & 26.99          & 0.26686               \\
\multicolumn{2}{c|}{E3D-LSTM\cite{e3dlstm}}& 1442.5          & 0.9803          & 32.52 & 0.04133                  & 1946.2          & 0.9047          & 25.45          & 0.12602      & 892.7          & 0.8153          & 21.78          & 0.48358      \\
\multicolumn{2}{c|}{PredNet\cite{prednet}}& 1625.3          & 0.9786          & 31.76 & 0.03264                  & 1568.9          & 0.9286          & 27.21          & 0.11289     & 783.1          & 0.8094          & 22.45          & 0.32159       \\
\multicolumn{2}{c|}{PhyDNet\cite{phydnet}}& 1614.7          & 0.9804          & \textbf{39.84} & 0.03709             & 2754.8          & 0.8615          & 23.26          & 0.32194   & 765.6          & 0.8322          & 23.41          & 0.50155              \\
\multicolumn{2}{c|}{MAU\cite{mau}}& 1577.0          & 0.9812          & 33.33 & 0.03561          & 1800.4          & 0.9176          & 26.14          & 0.09673    & 471.2          & 0.8945          & 26.73          & 0.25442                \\
\multicolumn{2}{c|}{MIM\cite{mim}}& 1467.1          & 0.9829          & 33.97 & 0.03338         & 1464.0          & 0.9409          & 28.10          & 0.06353     & 380.8          & 0.9025          & 27.78          & 0.18808   \\
\multicolumn{2}{c|}{PredRNN\cite{predrnn}}& 1458.3          & 0.9831          & 33.94 & 0.03245                  & 1525.5          & 0.9374          & 27.81          & 0.07395      & 380.6          & 0.9097          & 27.95          & 0.21892      \\
\multicolumn{2}{c|}{PredRNN++\cite{predrnn++}}& 1452.2          & 0.9832          & 34.02 & 0.03196          & 1453.2          & 0.9433          & 28.02          & 0.13210     & 370.4          & \textbf{0.9124} & \textbf{28.13} & \underline{0.19871}         \\
\multicolumn{2}{c|}{PredRNN.V2\cite{predrnnv2}}& 1484.7          & 0.9827          & 33.84 & 0.03334          & 1610.5          & 0.9330          & 27.12          & 0.08920      & 368.8          & 0.9099          & \underline{28.01}    & 0.21478              \\
\multicolumn{2}{c|}{TAU\cite{tau}}& 1390.7          & \underline{0.9839}  & 34.03 & 0.02783              & 1507.8          & 0.9456          & 27.83          & \underline{0.05494}  & 421.7          & 0.9086          & 27.10          & 0.22856        \\
\multicolumn{2}{c|}{DMVFN~\cite{dmvfn}}& -   & -  & - & -             & 1531.1          & 0.9314          & 26.95          & \textbf{0.04942}  & 413.2          & 0.8976          & 26.65          & \textbf{0.12842}        \\
\multicolumn{2}{c|}{SimVP (w/o VQ)\cite{simvpv2}}& 1441.0          & 0.9834          & \underline{34.08} & 0.03224           & 1507.7          & 0.9453          & 27.89          & 0.05740      & 397.1          & 0.9065          & 27.46          & 0.26496             \\
\multicolumn{2}{c|}{\textbf{SimVP+SVQ (Frozen codebook)}}& \textbf{1264.9} & \textbf{0.9851} & 34.07 & \underline{0.02380}   & \textbf{1408.6} & \textbf{0.9469} & \textbf{28.10} & 0.05535 & \underline{364.6}    & 0.9109          & 27.28          & 0.20988        \\
\multicolumn{2}{c|}{\textbf{SimVP+SVQ (Learnable codebook)}}& \underline{1265.1} & \textbf{0.9851}   & 34.06 &\textbf{0.02367}    & \underline{1414.9}    & \underline{0.9458}    & \textbf{28.10} & 0.05776& \textbf{360.2} & \underline{0.9116}    & 27.37          & 0.20658        \\
\multicolumn{2}{c|}{\textbf{Improvement}}& $\uparrow$\textbf{12.2\%}          & $\uparrow$\textbf{0.2\%}          &     $\downarrow$0.0\%    & $\uparrow$\textbf{26.2\%} &$\uparrow$\textbf{6.6\%}          & $\uparrow$\textbf{0.2\%}         & $\uparrow$\textbf{0.8\%}          & $\uparrow$\textbf{3.6\%}     &$\uparrow$\textbf{9.3\%}          & $\uparrow$\textbf{0.6\%}          & $\downarrow$0.3\%          & $\uparrow$\textbf{22.0\%}     \\

\bottomrule

\end{tabular}
\label{tab:bench_videos}
}
% \end{sc}
\end{small}
\end{center}
\end{table*}

% We assessed our model against recent state-of-the-art baselines, detailed in Table \ref{tab:weatherbench} and Table \ref{tab:bench_videos}, spanning non-recurrent and recurrent architectures. 

The benchmark results of WeatherBench, TaxiBJ and three video prediction datasets (Human3.6M, KTH, and KittiCaltech) are presented in Tables \ref{tab:weatherbench}, \ref{tab:taxibj} and \ref{tab:bench_videos}, respectively. These datasets have different characteristics. WeatherBench and TaxiBJ are macro forecasting tasks with low-frequency collection (30min or 1-6h). Human3.6M features subtle, low-frequency frame differences. KittiCaltech is challenging due to rapidly changing backgrounds and limited training data. The KTH dataset tests long-horizon forecasting, requiring the prediction of 20 future frames from 10 observed frames. 

Despite the distinct characteristics of each dataset, a consistent requirement across tasks is the need for effective noise reduction and pattern reservation, which universally benefit forecasting tasks. Notably, the SimVP+SVQ model achieves either the best or comparable performance across all datasets. For instance, on the WeatherBench-S temperature dataset, SVQ improves the best baseline by \textbf{7.9\%} (1.105 → 1.018). On three popular video prediction tasks, SVQ not only reduces forecasting errors (average \textbf{9.4\%} decrease in MAE), but also significantly improves subjective image quality (average \textbf{17.3\%} decrease in LPIPS).

\begin{figure*}
  \centering
  \includegraphics[width=1\linewidth]{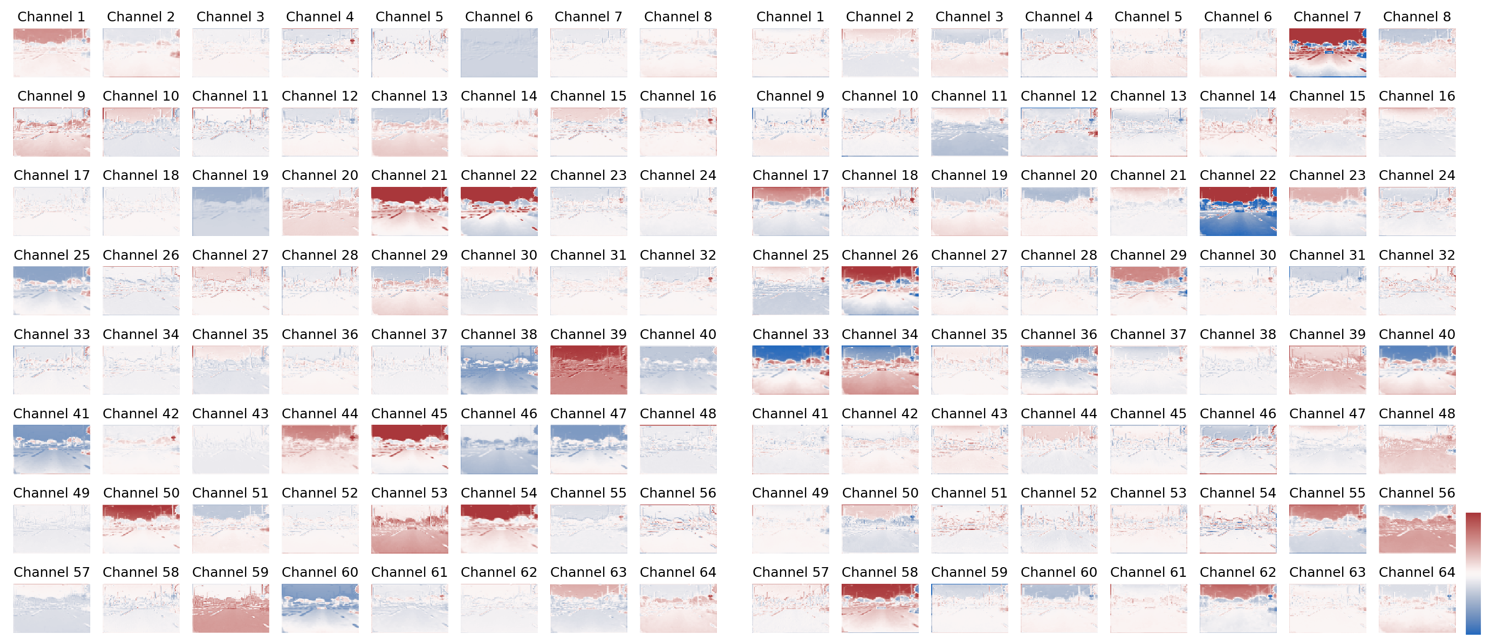}
  \vskip -0.1in
  \caption{Latent feature maps on the KittiCaltech dataset: Comparison before (Left) and after (Right) applying SVQ.}
  \label{fig:featuremap_kitti}
\end{figure*}

Besides, we explore both fixed (frozen) and learnable versions of the SVQ codebook on these forecasting tasks. Interestingly, our findings reveal that with a large codebook size, the performance of a frozen, randomly-initialized codebook is on par with that of a carefully learned codebook. This observation aligns with our intuition: when allowed to choose a very large number of representative vectors to form a codebook, due to high orthogonality, a random choice is often as good as the one that is carefully chosen. Similar phenomena have been studied in the context of the column subset selection problem in matrix theory~\cite{DBLP:journals/siammax/DrineasMM08,DBLP:conf/focs/DeshpandeR10}.

To further investigate the impact of SVQ on latent representations, we compared the feature maps before and after applying SVQ. Using the KittiCaltech dataset as an example, Figure~\ref{fig:featuremap_kitti} demonstrates that the difference between foreground and background becomes more obvious after applying SVQ. In particular, road conditions and the sky are more clearly separated. These results suggest that SVQ effectively extracts meaningful patterns and enhances the discriminative capacity of latent representations, which in turn improves downstream forecasting performance.

\subsection{Boosting performance as a versatile plug-in}

In this section, SVQ serves as a versatile plug-in module applicable to various MetaFormers~\cite{metaformer}. The adopted MetaFormers include three types. 1) CNN-based: SimVPv1(IncepU)~\cite{SimVP}, SimVPv2(gSTA)~\cite{simvpv2}, ConvMixer~\cite{convmixer}, ConvNeXt~\cite{convnext}, HorNet~\cite{hornet}, and MogaNet~\cite{moganet}. 2) Transformer-based: ViT~\cite{ViT}, Swin Transformer~\cite{SwinTransformer}, Uniformer~\cite{uniformer}, Poolformer~\cite{poolformer}, and VAN~\cite{van}. 3) MLP-based: MLPMixer~\cite{mlpmixer}. We conduct experiments on WeatherBench-S temperature dataset because it is lightweight and fast for training.

As shown in Table \ref{tab:boost}, SVQ consistently improves the performance across all MetaFormers, showcasing its universality across diverse backbone types. We observe an average reduction in MSE and MAE by \textbf{4.8\%} and \textbf{6.0\%}, respectively. In detail, SVQ leads to an average MSE reduction of 4.1\% for CNN-based backbones, 5.1\% for transformer-based, and 10.7\% for MLP-based. The more pronounced enhancement in transformer-based and MLP-based models indicates that our approach is especially effective with architectures that prioritize global interactions. Notably, SimVPv2(gSTA) is the best backbone, while our SVQ further improves it by 7.9\%. These findings also aligns with our motivation that mitigating noise in the learning process significantly benefits spatio-temporal forecasting, irrespective of model architecture. By integrating SVQ to constrain the diversity of predicted patterns and cut out noise, researchers can focus on crafting high-quality and general base models.
% \vskip -0.1in

\begin{table}[h]
\caption{{\bf Boosting performance:} The effect of SVQ for various MetaFormers on WeatherBench-S temperature dataset.}
\vskip -0.1in
\scalebox{0.9}{
\begin{tabular}{c|cc|cc}
\toprule
\multirow{2}{*}{MetaFormer} & \multicolumn{2}{c|}{\textbf{MSE}} & \multicolumn{2}{c}{\textbf{MAE}}\\
\cmidrule{2-5}
& w/o SVQ & w SVQ& w/o SVQ & w SVQ \\
\midrule
SimVPv1(IncepU)\cite{SimVP} & 1.238   & \textbf{1.216}      & 0.7037  & \textbf{0.6831}    \\
SimVPv2(gSTA)\cite{simvpv2}   & 1.105   & \textbf{1.018} & 0.6567  & \textbf{0.6109} \\
ConvMixer\cite{convmixer}   & 1.267   & \textbf{1.257}   & 0.7073  & \textbf{0.6780}   \\
ConvNeXt\cite{convnext}   & 1.277   & \textbf{1.159}    & 0.7220  & \textbf{0.6568}  \\
HorNet\cite{hornet}  & 1.201   & \textbf{1.130}    & 0.6906  & \textbf{0.6472}   \\
MogaNet\cite{moganet}     & 1.152   & \textbf{1.067}   & 0.6665  & \textbf{0.6271}  \\
\midrule

ViT\cite{ViT} & 1.146   & \textbf{1.111} & 0.6712  & \textbf{0.6375}\\
Swin\cite{SwinTransformer}                 & 1.143   & \textbf{1.088}  & 0.6735  & \textbf{0.6320} \\
Uniformer\cite{uniformer}            & 1.204   & \textbf{1.110}     & 0.6885  & \textbf{0.6400}     \\
Poolformer\cite{poolformer}  & 1.156   & \textbf{1.097} & 0.6715  & \textbf{0.6297} \\
VAN\cite{van}     & 1.150   & \textbf{1.083}     & 0.6803  & \textbf{0.6342} \\
\midrule

MLP-Mixer\cite{mlpmixer}   & 1.255   & \textbf{1.120}   & 0.7011  & \textbf{0.6455}   \\

\midrule
\textbf{Average improvement} & \multicolumn{2}{c|}{\textbf{$\uparrow$ 4.8\%}}  & \multicolumn{2}{c}{\textbf{$\uparrow$ 6.0\%}} \\
\bottomrule
\end{tabular}
}
\label{tab:boost}
\end{table}

\subsection{Delicate balance between detail preservation and noise reduction}

\begin{table}
\caption{{\bf Comparison of VQ methods on WeatherBench-S temperature dataset:} All methods share identical backbone, with the recommended setting in Appendix \ref{sec:appendix_vq_parameter}. The results better than baseline are highlighted in \textbf{bold}.}
\begin{center}
\vskip -0.1in
\begin{small}
% \begin{sc}
\scalebox{0.95}{
\begin{tabular}{c|cc}
\toprule
Method &MSE$\downarrow$ & MAE$\downarrow$ \\
\midrule
Baseline (SimVP w/o VQ)& 1.105 & 0.6567\\
\midrule
VQ~\cite{vqvae}& 1.854 & 0.8963 \\
Residual VQ (RVQ)~\cite{residualvq}& 1.213 & 0.6910 \\
Grouped Residual VQ (GRVQ)~\cite{groupresidualvq}& 1.174 & 0.6747 \\
Multi-headed VQ (MHVQ)~\cite{multiheadvq}&1.211 &0.6994 \\
Stochastic Residual VQ (SRVQ)~\cite{residualvq-stochastic}& 1.888 & 0.9237 \\
Residual Finite Scalar Quantization (RFSQ)~\cite{fsq}& 1.319 & 0.7505 \\
Lookup Free Quantization (LFQ)~\cite{lookupfreevq} &	2.988 & 1.1103\\
Residual LFQ (RLFQ)~\cite{lookupfreevq}& 1.281 & 0.7281\\
SVQ-raw~\cite{scvae}& 1.123 & \textbf{0.6456}\\
SVQ& \textbf{1.018} &	\textbf{0.6109}\\
\bottomrule
\end{tabular}
}
\label{tab:ablation_vq_method}
% \end{sc}
\end{small}
\end{center}
\vskip -0.15in
\end{table}

% \begin{table}[H]
% \centering
% % \captionsetup{font=small} 
% \vskip -0.15in
% \caption{\small  {\bf Comparison of vector quantization methods:} All methods share identical backbone, with the recommended setting as described in Appendix \ref{sec:appendix_vq_parameter}. The performance enhancements over baseline are highlighted in bold.}
% \begin{center}
% \begin{small}
% % \begin{sc}
% \scalebox{0.8}{

% \begin{tabular}{c|cc}
% \toprule
% Method &MSE$\downarrow$ & MAE$\downarrow$ \\
% \midrule
% Baseline (SimVP w/o VQ)& 1.105 & 0.6567\\
% \midrule
% VQ\cite{vqvae}& 1.854 & 0.8963 \\
% Residual VQ\cite{residualvq}& 1.213 & 0.6910 \\
% Grouped Residual VQ\cite{groupresidualvq}& 1.174 & 0.6747 \\
% Multi-headed VQ\cite{multiheadvq}&1.211 &0.6994 \\
% Residual VQ (Stochastic)\cite{residualvq-stochastic}& 1.888 & 0.9237 \\
% Residual Finite Scalar Quantization\cite{fsq}& 1.319 & 0.7505 \\
% Lookup Free Quantization (LFQ)\cite{lookupfreevq} &	2.988 & 1.1103\\
% Residual LFQ\cite{lookupfreevq}& 1.281 & 0.7281\\
% SVQ-raw\cite{scvae}& 1.123 & \textbf{0.6456}\\
% SVQ& \textbf{1.018} &	\textbf{0.6109}\\
% \bottomrule
% \end{tabular}
% }
% \label{tab:ablation_vq_method}
% % \end{sc}
% \end{small}
% \end{center}
% \vskip -0.3in
% %\vspace{-5mm}
% \end{table}
% \begin{wraptable}{r}{7cm}
% \vskip -0.16in

% \end{wraptable}

To investigate the role of VQ in spatio-temporal forecasting, we evaluated several state-of-the-art VQ methods that are structurally comparable to SVQ, each implemented as a plug-in module integrated with the backbone forecasting model. Table \ref{tab:ablation_vq_method} shows that SVQ significantly improves forecasting performance, whereas other VQ methods lead to increased prediction errors. Enhanced detail retention within the representational capacity is associated with lower forecasting errors. In particular, traditional VQ suffers from notable information losses, as evidenced by a higher MSE of 1.854. In contrast, residual VQ and grouped residual VQ yield lower MSEs of 1.213 and 1.174, respectively, affirming their ability to preserve fine-grained information through recursive quantization. 

\begin{figure}[h]
\centering
%\vskip -0.2in
\includegraphics[width=0.4\textwidth]{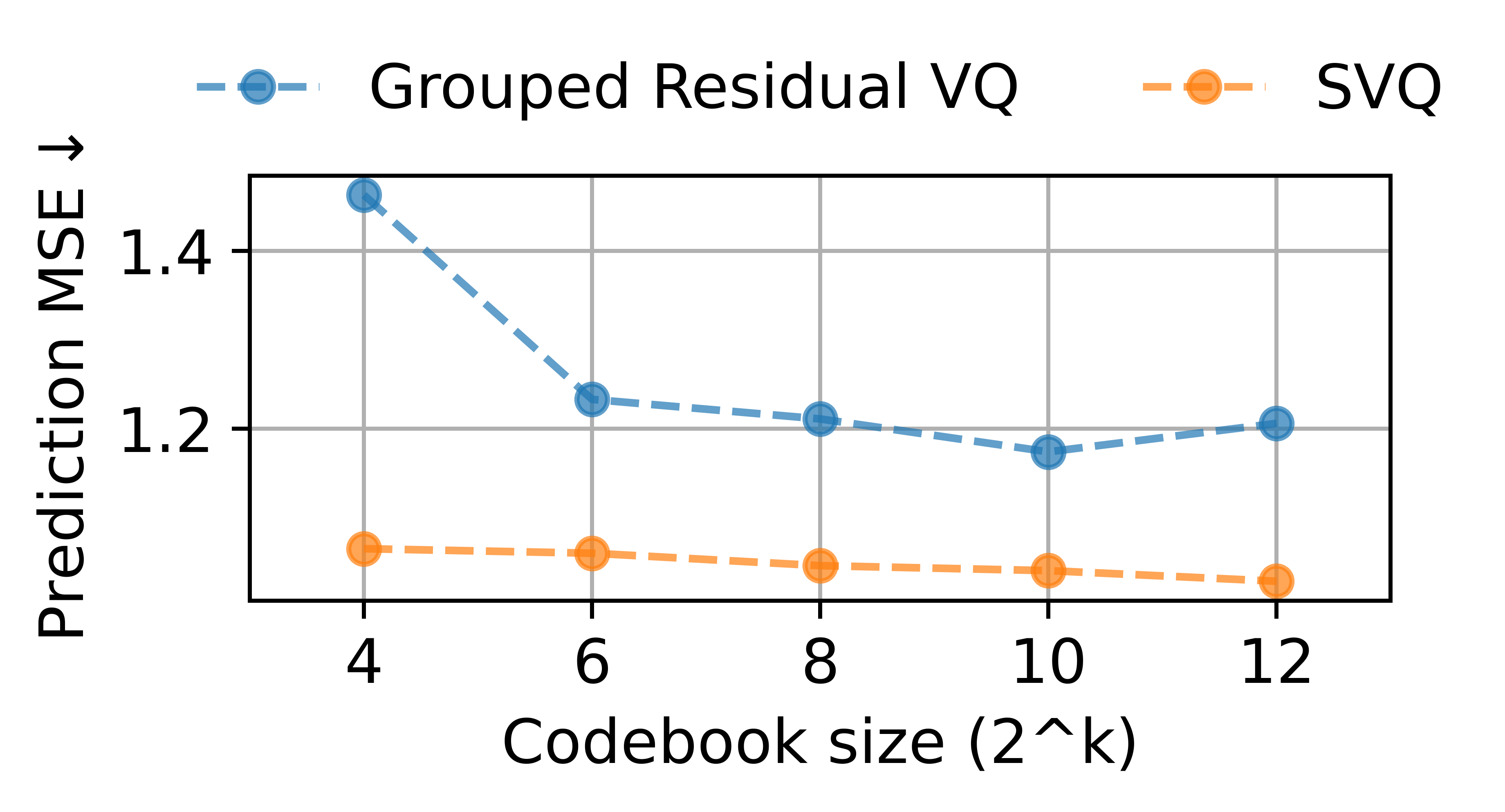}
\vskip -0.1in
\caption{{\bf Predition MSE curves} on WeatherBench-S temperature dataset with Grouped Residual VQ (GRVQ) and SVQ.}
\label{fig:compare_balance_vq}
\end{figure}

It is commonly understood that the codebook size in clustering-based VQ is critical: larger codebooks capture more details, whereas smaller ones enhance noise reduction. To explore this trade-off, we compared how the codebook size influences the prediction MSE in Grouped Residual VQ (GRVQ) and SVQ. As Figure \ref{fig:compare_balance_vq} indicates, the MSE of GRVQ initially decreases but increases with overly large codebooks, echoing findings from \cite{magvit2} that an excessively large codebook may degrade image generation performance. This underscores the necessity for dataset-specific tuning in clustering-based VQ approaches. In contrast, SVQ naturally achieves a balance between preserving detail and reducing noise through sparse regression, eliminating the need for extensive fine-tuning. The codebook size in SVQ exhibits a low-maintenance profile: using a default large codebook can produce robust results without extensive tuning. We are not suggesting that our SVQ outperforms other VQ methods in general image generation tasks, as it is beyond the scope of our current objective. Rather, we emphasize SVQ's effectiveness as a noise reduction tool that directly enhances real-world spatio-temporal forecasting tasks, while the application to general image generation remains a topic for future exploration.

% This outcome aligns with Section \ref{sec:comp_svq_cluster} that SVQ possesses better representation power compared to clustering-based quantization methods. We hypothesize that the regression-based quantization module endowed with strong representational capacity can directly improve forecasting performance. 

\subsection{Ablation study}
\label{sec:ablation}
We conduct a series of ablation studies on WeatherBench-S temperature dataset to understand the contribution of key design components based on the default setting: SVQ with a codebook size of 10,000, learnable codebook, and MAE loss.

\begin{figure*}
\centering
\scalebox{1}{
\includegraphics[width=0.85\linewidth]{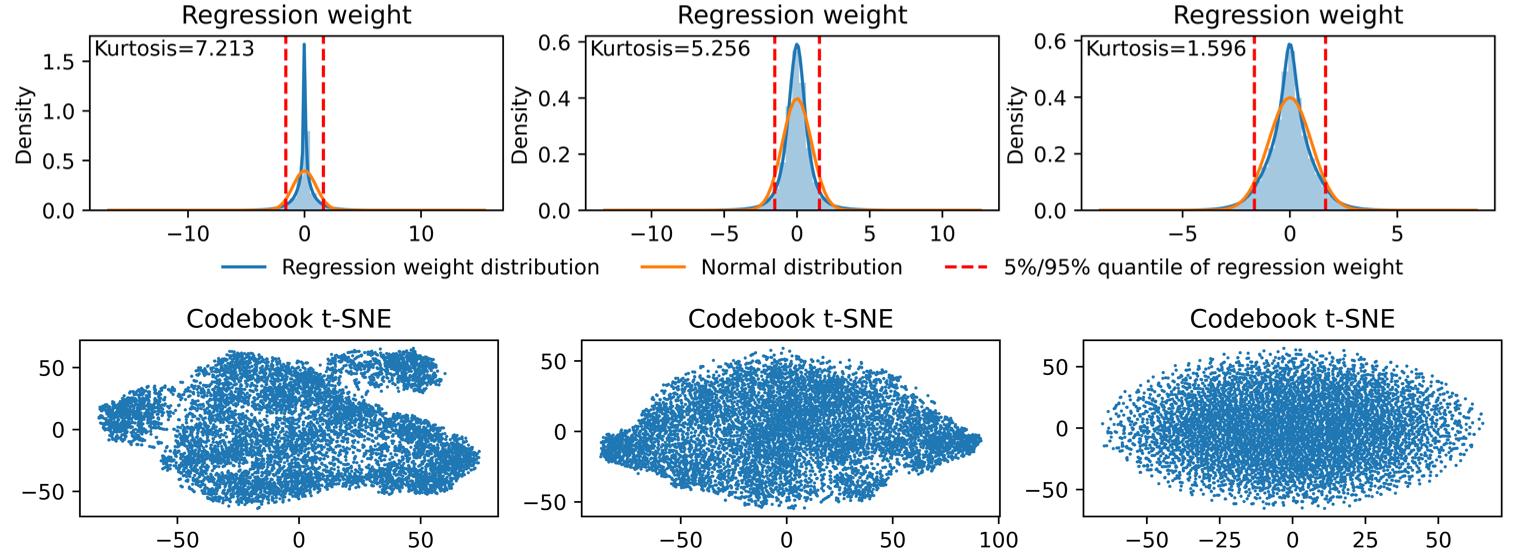}
}
\vskip -0.1in
\caption{{\bf Distribution of regression weight $\W$ and codebook $\M$:} Higher kurtosis represents more compact and concentrate distribution near zero, as well as sparser regression weights. Left: Learnable SVQ with MAE loss. Middle: Learnable SVQ with MSE loss. Right: Frozen SVQ with MAE loss. Learnable setting and MAE loss encourage sparser weights and a more structured codebook.}
\label{fig:sparsity}
\end{figure*}

\textit{Self-learned sparse regression structure.}\quad The original SimVP model adopts MSE as prediction loss. We individually replace it with MAE loss and add the SVQ module. As shown in Table \ref{tab:ablation_module}, the joint use of SVQ and MAE loss is crucial for significantly improving the model's performance. We suggest that the sparsity of the weight matrix $\W$ impacts vector representation learning and use kurtosis to quantify this after normalizing $\W$. Figure \ref{fig:sparsity} demonstrates that both a learnable codebook and MAE loss contribute to increased sparsity. Analyzing four codebook initialization methods in both learnable and fixed settings (Table \ref{tab:ablation_initialization}), we find that a learnable codebook promotes sparsity irrespective of the initialization, indicating that sparsity is a self-learned property that enhances intermediate representation learning.

\textit{Codebook size and learnability.}\quad Table \ref{tab:ablation_size_learnable} compares the effects of codebook size—both learnable and frozen. Results show that increasing codebook size consistently enhances performance. However, when the size reaches 10,000, the performance gap between frozen and learnable codebooks narrows to just 0.5\% (1.023 → 1.018). A larger codebook provides comprehensive coverage of the latent space through random codes, minimizing the need for meticulous learning. Consequently, models with randomly initialized codebooks perform similarly to those with learned ones. Additionally, our optimized SVQ structure outperforms alternative designs, including single-layer, bucket-shaped, and post-ReLU variants, while keeping a similar parameter count.

% \begin{table}
% %\vskip -0.2in
%   \centering
%   \begin{minipage}{0.48\textwidth} % Adjust width to fit your needs
%     \centering
%     \input{tables/tab_ablation_module.tex}
%   \end{minipage}
%   \hfill
%   %\hfill % Optional: this will add space between the two tables
%   \begin{minipage}{0.48\textwidth} % Adjust width to fit your needs
%     \centering
%     \input{tables/tab_ablation_initialization.tex}
%   \end{minipage}
% \end{table}

\begin{table}
\caption{{\bf Ablation of SVQ Compoments.}}
\begin{center}
\vskip -0.1in
% \begin{sc}
\scalebox{1.0}{
\begin{tabular}{c|cc}
\toprule
Module &MSE$\downarrow$ & MAE$\downarrow$ \\
\midrule
SimVP (MSE loss)& 1.105 & 0.6567  \\
SimVP (MAE loss)&	1.126 &	0.6509 \\
SimVP+SVQ (Learnable, MSE loss)&	1.099 &	0.6527 \\
SimVP+SVQ (Learnable, MAE loss)&	\textbf{1.018} &	\textbf{0.6109} \\
\bottomrule
\end{tabular}
\label{tab:ablation_module}
}
\end{center}
\end{table}

\begin{table}
\centering
% \captionsetup{font=small} 
\caption{{\bf Ablation of Codebook Initianlization.}}
\begin{center}
\begin{small}
\vskip -0.1in
% \begin{sc}
\scalebox{1.0}{
\begin{tabular}{c|c|ccc}
\toprule
Initialization & Learnability & MSE$\downarrow$ & MAE$\downarrow$ & Kurtosis\\
\midrule
\multirow{2}{*}{kaiming uniform} & Frozen & 1.023 & 0.6131 & 1.596 \\
&Learnable &\textbf{1.018} & \textbf{0.6109} & 7.213\\
\midrule
\multirow{2}{*}{sparse(sparsity=0.9)}&Frozen &1.050&0.6183 &4.165 \\
&Learnable & 1.034 &0.6160 &41.558\\
\midrule
\multirow{2}{*}{trunc normal}&Frozen &1.049 & 0.6166 &1.582 \\
&Learnable & 1.031 &0.6161 & 4.236\\
\midrule
\multirow{2}{*}{orthogonal}&Frozen &1.034&0.6170 &1.561 \\
&Learnable & 1.030 &0.6131 & 35.774\\
\bottomrule
\end{tabular}
% \end{sc}
}
\label{tab:ablation_initialization}
\end{small}
\end{center}
%\vspace{-5mm}
\end{table}

\begin{table}
\caption{{\bf Ablation of Model Structure.}}
\begin{center}
\vskip -0.1in
\begin{small}
% \begin{sc}
\scalebox{1.0}{
\begin{tabular}{c|c|c|cc}
\toprule
Learnability & Projection dim&Codebook size &	MSE$\downarrow$ & MAE$\downarrow$\\
\midrule
\multirow{3}{*}{Frozen} &128&10 & 1.070 & 0.6227 \\
&128 &1000& 1.044 & 0.6198 \\
&128 &10000& 1.023 & 0.6131 \\
\midrule
\multirow{5}{*}{Learnable}&128 &10 & 1.060 & 0.6194 \\
&128&1000 & 1.048 & 0.6182 \\
&128&10000 & \textbf{1.018} &	\textbf{0.6109}\\
\cmidrule{2-5}
&1280 (Bucket-shape) &1280 & 1.035 & 0.6149 \\
&None (One-layer) &10000 & 1.043 & 0.6144 \\
&128 (Post-ReLU) &10000 & 1.032 & 0.6136 \\
\bottomrule
\end{tabular}
\label{tab:ablation_size_learnable}
}
% \end{sc}
\end{small}
\end{center}
\end{table}
% \vskip -0.2in
%\vspace{-5mm}
% \end{table}

\textit{Frozen module.}\quad The SVQ module consists of a two-layer MLP and a large codebook. The MLP can be seen as a projection from the input vector to the regression weights. We have already examined the impact of freezing the codebook on forecasting performance. To further investigate the effect of freezing the MLP, we conducted an ablation study in this section. The results, presented in Table \ref{tab:frozenproj}, show that freezing the codebook only has a slight impact on forecasting performance, while freezing the MLP significantly impairs the performance. This highlights the critical role of the MLP in sparse regression, as it must be learned from data to generate the weights needed to combine codes from the codebook.
\begin{table}
\centering
% \captionsetup{font=small} 
\caption{Ablation of frozen modules.}
\begin{center}
\vskip -0.1in
% \begin{sc}
\scalebox{1.0}{
\begin{tabular}{c|cccc}
\toprule
\multirow{2}{*}{Metric} & \multicolumn{4}{c}{Frozen module} \\
\cmidrule{2-5}
 & None (All learnable) & Codebook & MLP & Both   \\
\midrule
MSE           & \textbf{1.018}      & 1.023   & 1.060                   & 1.093 \\
MAE           & \textbf{0.6109}      & 0.6131   & 0.6194                   & 0.6387 \\
\bottomrule
\end{tabular}
\label{tab:frozenproj}
}
% \end{sc}
\end{center}
%\vspace{-5mm}
\end{table}

\subsection{Train Stability Issue}
\label{sec:tability}

\begin{figure}[h]
  \centering
  \includegraphics[width=0.9\linewidth]{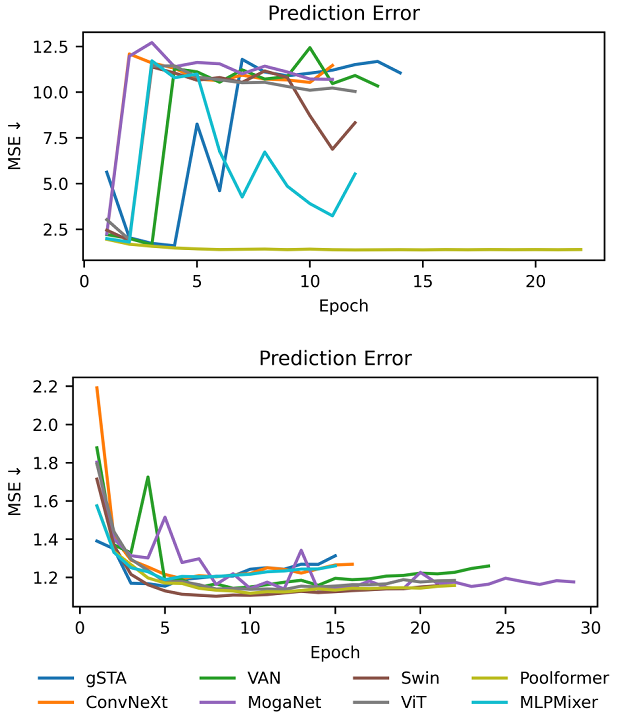}
  \caption{\textbf{VQ (Top) and SVQ (Bottom) training curves:} We perform post-translator quantization on various backbones, with the same codebook size (1024) on the WeatherBench-S temperature dataset.}
  \label{fig:vq_location}
\end{figure}

Although it is feasible to place the quantization module either before or after the translator, we found that for post-translator placement, the traditional VQ method~\cite{vqvae} suffers pronounced instability and codebook collapse issues, as shown in Figure \ref{fig:vq_location}. It is essential to highlight that the backbones without VQ maintain their MSE within the acceptable range of approximately 1 to 2 (refer to Table \ref{tab:boost}). Yet, integrating traditional VQ causes a substantial rise in MSE values, exceeding 10 for different backbones—a level considered excessively high. Since two designs only differ in the placement order of quantization module and translator module, we hypothesize that this instability is attributed to the non-differentiability of the straight-through estimator, which introduces errors into the gradient flow for preceding modules. In contrast, our SVQ module never encounters such issues and remains highly stable throughout training. To maintain the integrity of all VQ methods, we opt for the pre-translator design in our main experiments, wherein quantization is executed before the translator module.

% \subsection{Robustness to noise, error-bar, convergence behaviour, visualizations of predictions and latent vectors, and high-dimensional benchmark results.}

% We confirmed our method effectively mitigates noise by adding artificial noise to the training data and constraining latent patterns through quantization,as detailed in Appendix~\ref{sec:appendix_noise}. The computational cost introduced by SVQ is detailed in Appendix~\ref{sec:appendix_cost}. The statistical significance of the error bars is provided in Appendix~\ref{sec:statistical_significance}. We further analyzed the convergence behavior of SVQ and traditional VQ in Appendix~\ref{sec:convergence_analysis}. We show experiments to understand the impact of SVQ on latent representation and to compare various VQ methods in Appendix \ref{sec:appendix_latent}, and \ref{sec:additional_vq}, respectively.
% A benchmark experiment for high-dimensional forecasting is included in Appendix~\ref{sec:appendix_weatherbenchhmv}.
\section{Conclusions}
In this paper, we present Differentiable Sparse Soft-Vector Quantization (SVQ), an innovative and efficient technique for improving spatio-temporal forecasting. Unlike current state-of-the-art vector quantization methods, which often degrade performance, our approach is the first to show a beneficial impact on spatio-temporal forecasting tasks. SVQ elegantly tackles the inaccuracies in the optimization problem arising from non-differentiability and the restricted representational capabilities associated with hard-VQ. Tested across diverse benchmarks, from weather to traffic and video prediction, SVQ consistently outperforms pure baseline methods, setting new performance standards without complex prior models. Its differentiability and seamless integration with baseline models highlight SVQ as a significant advancement for efficient and effective spatio-temporal forecasting.

\bibliographystyle{ACM-Reference-Format}
\bibliography{bib,bib_film}

\appendix
% \onecolumn
% \begin{center}
% \textbf{\Large Supplemental Materials}
% \end{center}

% The supplementary material for our work  \textit{Does Vector Quantization Fail in Spatio-Temporal Forecasting? Exploring a Differentiable Sparse Soft-Vector Quantization Approach} is organized as follows: Appendix \ref{sec:appendix_implem} provides implementation details of SimVP model and VQ methods. Appendix \ref{additionalreview} gives an extensive review of related work. Appendix \ref{sec:convergence_analysis} analyzes the convergence behaviour of SVQ and traditional VQ. Appendix \ref{sec:appendix_exp} present extended quantitative results. Appendix \ref{sec:appendix_latent} delves deeper into the effect of SVQ on latent representation. Finally, Appendix \ref{sec:appendix_quatlitative} shows additional qualitative results of forecasting samples and errors. 

\section{Implementation details}
\label{sec:appendix_implem}
\subsection{Dataset details}
\label{sec:appendix_dataset}

% A summary of dataset statistics is provided in Table \ref{tab:dataset}.

%\vskip -0.1in
\begin{table}[H]
\centering
\caption{Statistics of benchmark datasets.}
\begin{center}
\vskip -0.1in
% \begin{small}
\scalebox{0.9}{
% \begin{sc}
\begin{tabular}{c|cc|cc|c|c}
\toprule
\multirow{2}{*}{Dataset} & \multicolumn{2}{c|}{Size} & \multicolumn{2}{c|}{Seq. Len.} & Img. Shape & \multirow{2}{*}{Interval}\\
& train & test & in & out & H $\times$ W $\times$ C &\\
\midrule
WeatherBench-S & 2,167 & 706 & 12 & 12 & 32 $\times$ 64 $\times$ 1& 1 hour\\
WeatherBench-M & 54,019 & 2,883 & 4 & 4 & 32 $\times$ 64 $\times$ 4 & 6 hour\\
TaxiBJ & 20,461 & 500 & 4 & 4 & 32 $\times$ 32 $\times$ 2& 30 min\\
KittiCaltech & 3,160 & 3,095 & 10 & 1 & 128 $\times$ 160 $\times$ 3&Frame\\
Human3.6M & 73,404 & 8,582 & 4 & 4 & 256 $\times$ 256 $\times$ 3&Frame\\
KTH Action & 4,940 & 3,030 & 10 & 20 & 128 $\times$ 128 $\times$ 1&Frame\\
% WeatherBench-HMV & 52,559 & 2,883 & 4 & 4 & 32 $\times$ 64 $\times$ 110 & 6 hour\\
\bottomrule
\end{tabular}
\label{tab:dataset}
}
% \end{sc}
% \end{small}
\end{center}
\vskip -0.1in
\end{table}

%\vskip -0.1in

\subsection{Architecture configuration of SimVP}
\label{sec:appendix_architecture}
Table \ref{tab:appendix_parameters} reports the architectures of SimVP on all datasets. We select the best MetaFormer variant to replace the translator module based on OpenSTL benchmarks\footnote{\url{https://openstl.readthedocs.io/en/latest/model_zoos/video_benchmarks.html}}\footnote{\url{https://openstl.readthedocs.io/en/latest/model_zoos/weather_benchmarks.html}}\footnote{\url{https://openstl.readthedocs.io/en/latest/model_zoos/traffic_benchmarks.html}}. The parameters remain unchanged, following the original configurations. It is noteworthy that due to reproducibility issues of ConvNeXt on the TaxiBJ dataset, we have opted to utilize gSTA as our backbone model.

\subsection{Parameters of compared VQ methods}
\label{sec:appendix_vq_parameter}
Table \ref{tab:ablation_vq_method} presents a comparison of SVQ with several well-known VQ methods, reproduced using source code from the GitHub repository\footnote{\url{https://github.com/lucidrains/vector-quantize-pytorch/tree/master}}. The parameters were kept consistent with the recommended settings to ensure performance, as detailed in Table \ref{tab:appendix_vqparameters}. It should be noted that we found that increasing the codebook size for traditional VQ methods, such as Residual VQ and Multi-headed VQ, led to a considerable increase in GPU memory usage and extended the training time to impractical levels. This issue is one of the reasons these methods recommend adopting a default codebook size of 1024. To ensure fairness, we conducted an extensive experiment for VQ methods using the same codebook size (1024) in Appendix \ref{sec:additional_vq}.

\section{Additional quantitative results}
\label{sec:appendix_exp}

\subsection{Computational cost}
\label{sec:appendix_cost}
Table \ref{tab:appendix_nparams} presents the computational costs of SVQ module and forecasting models. It shows that recurrent-based models have significantly higher FLOPs requirements, while non-recurrent models are more efficient. The proposed SVQ module is not only effective but also lightweight: across all datasets, it introduces only a minor increase in both parameter count and FLOPs. Notably, the overall computational cost of SimVP+SVQ remains substantially lower than that of recurrent-based models.

More importantly, the observed additional computational cost introduced by SVQ is primarily determined by the codebook size and dimensionality, and remains independent of the choice of base model. Consequently, if SVQ were applied to more complex baselines (e.g., ConvLSTM or PredRNN), the relative increase in FLOPs would be negligible.

\subsection{Comprehensive comparison of VQ methods using the same codebook size}
\label{sec:additional_vq}
We extend Table \ref{tab:ablation_vq_method} by setting the codebook size to the same value (1024) for compared VQ methods. They are comprehensively evaluated from different aspects: downstream performance (prediction MSE), codebook usage (perplexity), and computational complexity (FLOPS, inference FPS, and training time per epoch). Unlike other VQ methods that rely on a single code, our SVQ generates multiple regression weights to merge several codes. To evaluate its perplexity, we first normalize the regression weights and then convert them into binary form using a threshold set at $\theta$ times the standard deviation, where $\theta$ serves as the threshold value. Two thresholds (2 and 3) are utilized to obtain reasonable perplexity. 

\begin{figure}[h]
  \centering
  \includegraphics[width=1.0\linewidth]{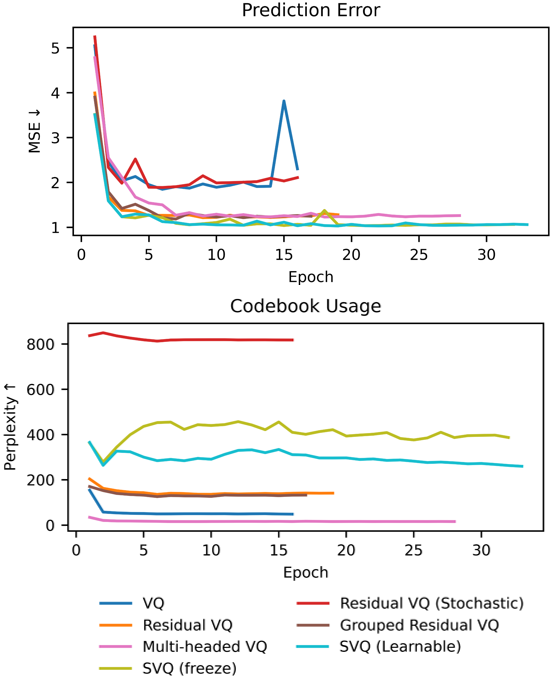}
  \caption{Prediction error and codebook usage of VQ methods during the training process. All methods adopt the same codebook size (1024) and are early stopped with a patience of 10, trained on WeatherBench-S temperature dataset. For simplicity, the perplexity of SVQ is averaged on different $\theta$.}
  \label{fig:vq_curve}
\end{figure}

The quantitative and convergence results are shown in Table \ref{tab:appendix_vqefficiency} and Figure \ref{fig:vq_curve}. SVQ quickly converges to the lowest prediction error and satisfactory utilization of the codebook. Residual VQ with stochastic sampling has the highest codebook usage. However, its prediction MSE is worse than residual VQ without stochastic sampling. This demonstrates that forcibly improving codebook usage does not guarantee better downstream performance. SVQ generally outperforms the other VQ methods in computational efficiency, due to the approximation described in Section \ref{sec:sparse_regression}.

\begin{table*}
\centering
% \captionsetup{font=small} 
\caption{Detailed configuration of SimVP backbone.}
\begin{center}
\vskip -0.1in
\begin{small}
% \begin{sc}
\scalebox{1.0}{
\begin{tabular}{c|ccccccccc}
\toprule
Dataset                          & MetaFormer (Translator) & spatio\_kernel & hid\_S & hid\_T & N\_T & N\_S & drop\_path & LR scheduler\\
\midrule
WeatherBench-S temperature       & gSTA       & enc=3, dec=3   & 32     & 256    & 8    & 2    & 0.1        & cosine        \\
WeatherBench-S humidity          & Swin       & enc=3, dec=3   & 32     & 256    & 8    & 2    & 0.2        & cosine       \\
WeatherBench-S wind component    & Swin       & enc=3, dec=3   & 32     & 256    & 8    & 2    & 0.2        & cosine       \\
WeatherBench-S total cloud cover & gSTA       & enc=3, dec=3   & 32     & 256    & 8    & 2    & 0.1        & cosine       \\
WeatherBench-M                   & MogaNet    & enc=3, dec=3   & 32     & 256    & 8    & 2    & 0.1        & cosine       \\
TaxiBJ                           & gSTA   & enc=3, dec=3   & 32     & 256    & 8    & 2    & 0.1        & cosine       \\
Human3.6M                        & gSTA       & enc=3, dec=3   & 64     & 512    & 6    & 4    & 0.1        & cosine       \\
KTH                              & IncepU     & enc=3, dec=3   & 64     & 256    & 6    & 2    & 0.1        & onecycle     \\
KittiCaltech                     & gSTA       & enc=3, dec=3   & 64     & 256    & 6    & 2    & 0.2        & onecycle     \\
% WeatherBench-HMV                   & gSTA    & enc=3, dec=3   & 32     & 256    & 8    & 2    & 0.1        & cosine       \\
\bottomrule
\end{tabular}
\label{tab:appendix_parameters}
% \end{sc}
}
\end{small}
\end{center}
%\vspace{-5mm}
\end{table*}

\begin{table*}
\centering
% \captionsetup{font=small} 
\caption{Parameters of the compared VQ methods.}
\begin{center}
\vskip -0.1in

% \begin{sc}
\scalebox{0.9}{
\begin{tabular}{c|cccccc}
\toprule
Vector quantization method          & codebook\_size & num\_quantizers & groups & heads & shared\_codebook & Specific parameters                                                                              \\
\midrule
VQ                                  & 512            & -               & -      & -     & -      & -     \\
Residual VQ                         & 1024           & 8               & -      & -    & $\checkmark$ & -\\
Grouped Residual VQ                 & 1024           & 8               & 2      & -    & $\checkmark$ & -\\
Multi-headed VQ                     & 1024           & -               & -      & 8   & $\checkmark$ & - \\
Residual VQ (Stochastic)    & 1024           & 8               & -      & -   & $\checkmark$ & stochastic\_sample\_codes=True\\
Residual Finite Scalar Quantization & -              & 8               & -      & -  & -     & levels={[}8, 5, 5, 3{]}                                                                          \\
Lookup Free Quantization (LFQ)      & 8192           & -               & -      & -   & -    & entropy\_loss\_weight=0.1\\
Residual LFQ                        & 256            & 8               & -      & -   & -    & -      \\
\bottomrule
\end{tabular}
\label{tab:appendix_vqparameters}
}
% \end{sc}
\end{center}
%\vskip -0.2in
\end{table*}

\begin{table*}
\centering
% \captionsetup{font=small} 
\caption{Number of parameters and computing performances for all forecasting models.}
\begin{center}
\vskip -0.1in
\begin{small}
% \begin{sc}
\scalebox{1.0}{
\begin{tabular}{c|c|cc|cc|cc|cc|cc}
\toprule
\multirow{2}{*}{Model type}      & Dataset        & \multicolumn{2}{c|}{Human3.6M} & \multicolumn{2}{c|}{KTH} & \multicolumn{2}{c|}{KittiCaltech} & \multicolumn{2}{c|}{WeatherBench-S} & \multicolumn{2}{c}{TaxiBJ} \\
\cmidrule{2-12}
                                 & Model          & Params        & FLOPs         & Params     & FLOPs      & Params          & FLOPs          & Params           & FLOPs           & Params       & FLOPs       \\
\midrule
\multirow{9}{*}{Recurrent-based} & ConvLSTM         & 15.5M           & 347.0G         & 14.9M           & 1368.0G        & 15.0M           & 595.0G         & 14.98M           & 136G            & 14.98M          & 20.74G         \\
                                 & E3D-LSTM         & 60.9M           & 542.0G         & 53.5M           & 217.0G         & 54.9M           & 1004G          & 51.09M           & 169G            & 50.99M          & 98.19G         \\
                                 & PredNet          & 12.5M           & 13.7G          & 12.5M           & 3.4G           & 12.5M           & 12.5M          & -                & -               & 12.5M           & 0.85G          \\
                                 & PhyDNet          & 4.2M            & 19.1G          & 3.1M            & 93.6G          & 3.1M            & 40.4G          & 3.09M            & 36.8G           & 3.09M           & 5.60G          \\
                                 & MAU              & 20.2M           & 105.0G         & 20.1M           & 399.0G         & 24.3M           & 172.0G         & 5.46M            & 39.6G           & 4.41M           & 6.02G          \\
                                 & MIM              & 47.6M           & 1051.0G        & 39.8M           & 1099.0G        & 49.2M           & 1858G          & 37.75M           & 109G            & 37.86M          & 64.10G         \\
                                 & PredRNN          & 24.6M           & 704.0G         & 23.6M           & 2800.0G        & 23.7M           & 1216G          & 23.57M           & 278G            & 23.66M          & 42.40G         \\
                                 & PredRNN++        & 39.3M           & 1033.0G        & 38.3M           & 4162.0G        & 38.5M           & 1803G          & 38.31M           & 413G            & 38.40M          & 62.95G         \\
                                 & PredRNN.V2       & 24.6M           & 708.0G         & 23.6M           & 2815.0G        & 23.8M           & 1223G          & 23.59M           & 279G            & 23.67M          & 42.63G         \\
                                 & DMVFN       & -           & -         & 3.5M           & 0.88G        & 3.6M           & 1.2G          & -           & -            & 3.54M          & 0.057G         \\
\midrule
\multirow{3}{*}{Non-recurrent}   & TAU              & 37.6M           & 182.0G         & 15.0M           & 73.8G          & 44.7M           & 80.0G          & 12.22M           & 6.70G           & 9.55M           & 2.49G          \\
                                 & SimVP (w/o VQ)   & 28.8M           & 146.0G          & 12.2M           & 62.8G          & 15.6M           & 96.3G          & 12.76M           & 7.01G           & 7.84M           & 2.08G          \\
                                 & \textbf{SimVP+SVQ}        & 30.7M           & 178.0G         & 13.3M           & 110.0G         & 16.8M           & 156G           & 14.37M           & 16.8G           & 9.45M           & 3.72G   \\
\bottomrule
\end{tabular}
\label{tab:appendix_nparams}
}
% \end{sc}
\end{small}
\end{center}
% \vspace{-5mm}
\end{table*}

\begin{table*}
\centering
% \captionsetup{font=small} 
\caption{Quantitative comparison of VQ methods with the same codebook size (1024) on WeatherBench-S temperature dataset.}
\begin{center}
\begin{small}
\scalebox{1.0}{
\begin{tabular}{c|ccccc}
\toprule
Vector quantization method & Prediction MSE$\downarrow$ & Perplexity$\uparrow$    & FLOPs$\downarrow$  & Inference FPS$\uparrow$ & Training time per epoch (min)$\downarrow$ \\
\midrule
VQ                         & 1.8544         & 51.95         & 7.207G & 21.1          & 7.11          \\
Residual VQ                & 1.2131         & 142.47        & 8.616G & 7.7           & 13.25         \\
Residual VQ (Stochastic)   & 1.8882         & 817.91        & 8.616G & 8.1           & 17.27         \\
Grouped Residual VQ        & 1.1737         & 132.57        & 8.616G & 4.8           & 19.98         \\
Multi-headed VQ            & 1.2113         & 16.36         & 8.717G & 6.2           & 13.15         \\
SVQ (Frozen)               & 1.0393         & 335.72($\theta$=3)/438.39($\theta$=2) & 8.037G & 24.6          & 7.27          \\
SVQ (Learnable)                       & 1.0403         & 246.44($\theta$=3)/331.41($\theta$=2) & 8.037G & 24.9          & 7.30    \\
\bottomrule
\end{tabular}
}
\end{small}
\end{center}
\label{tab:appendix_vqefficiency}
\end{table*}

\clearpage

\end{document}